\documentclass[10pt,a4paper,twocolumn]{article}
\usepackage[margin=18mm,columnsep=6mm]{geometry}

\usepackage[T1]{fontenc}
\usepackage[utf8]{inputenc}
\PassOptionsToPackage{table,dvipsnames}{xcolor}
\usepackage{amsmath}
\usepackage{amssymb}
\usepackage{txfonts}
\usepackage{bm}
\usepackage{microtype}
\usepackage{graphicx,booktabs,array,multirow}
\usepackage{xcolor}
\usepackage{caption}
\usepackage{placeins}
\usepackage{needspace}\usepackage{cuted}
\usepackage{enumitem}
\usepackage[numbers,sort&compress]{natbib}
\usepackage{hyperref}
\usepackage{ragged2e}
\hypersetup{colorlinks=true,linkcolor=MidnightBlue,citecolor=MidnightBlue,urlcolor=MidnightBlue}
\usepackage[capitalise,noabbrev]{cleveref}
\setlist[itemize]{leftmargin=*,topsep=3pt,itemsep=2pt}
\setlist[enumerate]{leftmargin=*,topsep=3pt,itemsep=2pt}
\definecolor{rankblueone}{RGB}{173,210,238}
\definecolor{rankbluetwo}{RGB}{206,228,245}
\definecolor{rankbluethree}{RGB}{232,242,250}
\newcommand{\rankone}[1]{\cellcolor{rankblueone}#1}
\newcommand{\ranktwo}[1]{\cellcolor{rankbluetwo}#1}
\newcommand{\rankthree}[1]{\cellcolor{rankbluethree}#1}
\newcommand{\method}{ClearForm}
\newcommand{\sg}{\operatorname{sg}}
\newcommand{\clip}{\operatorname{clip}}
\newcommand{\norm}[1]{\left\lVert #1\right\rVert}
\newcommand{\ind}{\mathbf{1}}
\newcommand{\papertitle}{ClearForm: Shared-Surface Gaussian Splatting\\for Transparent and Opaque Scenes}

\newcommand{\FrontCD}{27.05}

\title{\papertitle}

\author{
Hajeong Lee \qquad Jong-Il Park\textsuperscript{*}\\[4pt]
\normalsize Hanyang University, Seoul, Republic of Korea\\[2pt]
\normalsize
\texttt{hajeonglee25@hanyang.ac.kr}
\quad
\texttt{jipark@hanyang.ac.kr}\\[2pt]
\small \textsuperscript{*}Corresponding author
}
\date{}
\begin{document}
\makeatletter
\twocolumn[{\@maketitle
\vspace{-1.0em}
\begin{center}
\includegraphics[width=\textwidth]{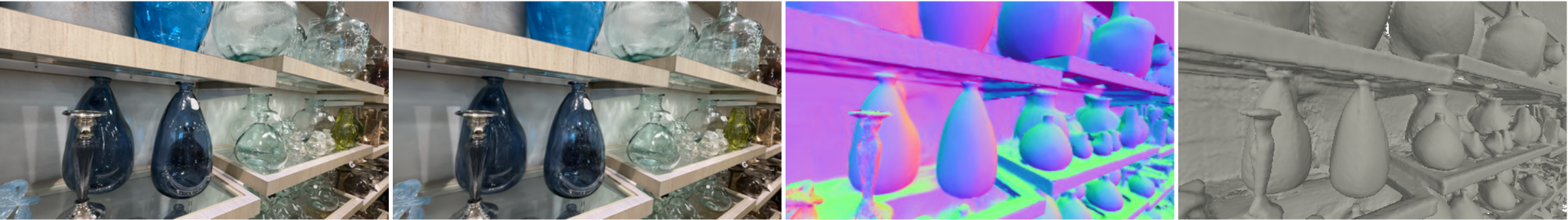}
\captionof{figure}{\textbf{ClearForm on Ref-DL3DV.} The rendering preserves the appearance of the transparent vessels while the reconstructed mesh shows their surfaces. From left to right: reference RGB, ClearForm rendering, normals, and reconstructed mesh. The scene is from the Ref-DL3DV subset of DL3DV-10K~\citep{na2026glint,ling2024dl3dv}.}
\label{fig:teaser}
\end{center}
\vspace{0.8em}}]
\makeatother
\begin{abstract}

% BEGIN abstract.tex
Transparent-interface reconstruction must recover glass surfaces even when most observed color comes from the scene behind them. We present \method{}, a shared-planar-Gaussian method for jointly reconstructing transparent interfaces and the appearance seen through them. We identify and analyze the \emph{Geometry--Appearance Compositing Dilemma} (GACD): when the same transparency-dependent alpha controls both color and geometry, revealing the background can weaken a transparent surface's contribution to geometric reconstruction. We derive its effect on plane-based depth estimation, identify a mechanism that can weaken geometric learning, and quantify the loss of interface support through controlled comparisons. To address this conflict, we develop \emph{geometry-preserving dual-opacity surface reconstruction}. Shared planar Gaussians retain interface contributions in geometry compositing while modeling transmission and sparse reflection in RGB, supported by transparency-aware geometric constraints. To manage the different primitive-visibility requirements of these compositors, we introduce \emph{Dual-Evidence Lifecycle} (DEL), which retains Gaussians supported by multi-view geometry or RGB reconstruction evidence. On 3D-FRONT-T, \method{} achieves 27.05\,mm Chamfer distance (CD) and 0.928 F1, compared with 59.18\,mm CD for our PGSR reproduction and the reported 34\,mm CD and 0.836 F1 for GLINT. On TransLab, it achieves 1.65\,mm CD, 0.963 F1, and 40.14\,dB PSNR. In matched ablations, dual-opacity compositing reduces CD from 59.25 to 27.09\,mm, while DEL improves PSNR by 2.06\,dB.
Our project page is available at
\url{https://hajeongx.github.io/clearform/}.

% END abstract.tex

\end{abstract}

% BEGIN paper.tex
\section{Introduction}
Glass transmits light while introducing a physical boundary between materials. We call this boundary a transparent interface. Along a camera ray, the near-side interface lies before the scene surfaces visible through the glass, which we call background geometry. The appearance of these background surfaces as seen through the interface constitutes transmitted appearance. Reconstruction must therefore recover interface geometry while preserving the observed appearance.

3D Gaussian Splatting (3DGS)~\citep{kerbl2023}
renders color through front-to-back alpha compositing,
where each Gaussian is weighted by its opacity and
the transmittance remaining after nearer Gaussians.
Surface-oriented methods improve geometry through
surface alignment, planar primitives, or opacity-field
extraction~\citep{guedon2024,huang2024,yu2024gof,chen2024pgsr}.
Planar-based Gaussian Splatting (PGSR)~\citep{chen2024pgsr}
also uses these compositing weights to blend plane
normals and distances for depth estimation.
When the same alpha is reduced to reveal background
appearance through a transparent interface, its
geometric contribution is reduced as well.

We call this conflict the \emph{Geometry--Appearance Compositing Dilemma} (GACD). For a single plane, its alpha weight cancels out in PGSR's depth calculation. However, when transparent interfaces and background surfaces contribute to the same pixel, transparency changes their relative weights and can shift the estimated depth. It can also weaken a geometric gradient pathway and cause small interface contributions to be skipped during rendering. We derive these effects and measure how shared alpha reduces support for the ground-truth first surface. Separate training ablations assess whether geometry-preserving compositing improves reconstruction accuracy.

Guided by this analysis, we develop \emph{geometry-preserving dual-opacity surface reconstruction} over shared planar Gaussians. The geometry path blends plane attributes without transparency attenuation, while the appearance path reveals background color through learned transparency and represents sparse reflection on the same surface. Transparency-aware depth--normal consistency and multi-view geometric constraints then supervise the interface without requiring it to block transmitted color in RGB. Together, these components reconstruct transparent-interface geometry, transmitted appearance, and reflection.

Dual-opacity compositing also separates the visibility conditions for geometry and RGB. A transparent interface can contribute strongly to geometric rendering while allowing background color to pass through. It may therefore occlude background Gaussians in the geometry pass even though they remain useful for transmitted RGB. If their learned opacity becomes small during optimization, conventional opacity-based pruning can remove them; a geometry-only visibility test likewise misses their appearance contribution. We introduce \emph{Dual-Evidence Lifecycle} (DEL), which retains Gaussians supported by geometry-pass visibility or positive RGB-loss contribution across distinct camera views, and maintains evidence across cloning and splitting.

At a fixed Gaussian state, changing only the geometry alpha rule raises ground-truth first-hit coverage from 7.08\% to 14.70\%. In the cumulative 3D-FRONT-T ablation, dual-opacity compositing reduces Chamfer distance (CD) from 59.25 to 27.09\,mm; DEL then raises PSNR from 32.05 to 34.11\,dB while retaining similar mesh accuracy. The complete model achieves 27.05\,mm CD and 0.928 F1 on 3D-FRONT-T, and 1.65\,mm CD and 0.963 F1 on TransLab.

Our contributions are summarized as follows:
\begin{itemize}
\item \textbf{Geometry--Appearance Compositing Dilemma.} We analyze how transparency-dependent alpha can weaken the geometric contributions of transparent interfaces while revealing the background, including its effects when different surfaces overlap along a ray. We derive the consequences for depth rendering and geometric optimization and quantify the loss of first-surface support. 
\item \textbf{Geometry-preserving dual-opacity surface reconstruction.} We formulate separate geometry and appearance compositing over shared planar Gaussians so transparent interfaces can contribute to surface reconstruction without obscuring transmitted color. Transparency-aware PGSR regularization constrains interface geometry, while learned transparency and sparse reflection reproduce the observed appearance.
\item \textbf{Dual-Evidence Lifecycle.} We introduce a multi-view Gaussian survival criterion that addresses the geometry--RGB visibility mismatch in dual-opacity reconstruction. By retaining primitives supported by geometry-pass visibility or positive RGB-loss contribution across distinct cameras, DEL helps preserve transparent interfaces and appearance-useful background Gaussians with different rendering roles.
\end{itemize}

\section{Related Work}
\paragraph{Surface reconstruction with neural and Gaussian representations.}
Neural surface methods including VolSDF, NeuS, and Neuralangelo~\citep{yariv2021,wang2021neus,li2023neuralangelo} extend radiance-field optimization~\citep{mildenhall2020} with surface structure or geometric regularization. Gaussian methods offer efficient alternatives: SuGaR aligns Gaussians to surfaces~\citep{guedon2024}, 2DGS uses oriented disks~\citep{huang2024}, Gaussian Opacity Fields extracts surfaces from opacity~\citep{yu2024gof}, and PGSR estimates depth from alpha-blended plane normals and distances while enforcing single- and multi-view constraints~\citep{chen2024pgsr}. GS-2DGS adds learned geometric guidance for reflective objects~\citep{tong2025gs2dgs}. In ClearForm, PGSR's plane representation provides the surface attributes for a distinct geometry-compositing path.

\paragraph{Reflective and refractive appearance.}
Ref-NeRF, NeRFReN, and NeRO address non-Lambertian appearance through directional representations, radiance decomposition, and geometry/material estimation~\citep{verbin2022,guo2022nerfren,liu2023nero}. GlassGaussian models glass reflectance~\citep{cao2025glassgaussian}; Ref-GS and EnvGS capture view-dependent reflections~\citep{zhang2025refgs,xie2025envgs}, and GUS-IR combines forward and deferred shading~\citep{liang2025gusir}. Other methods model refractive transport: Eikonal Fields traces curved rays~\citep{bemana2022}, NU-NeRF reconstructs nested transparent objects~\citep{sun2024nunerf}, and TransparentGS uses deferred refraction~\citep{huang2025transparentgs}. ClearForm instead uses straight-ray splatting and learned view-dependent appearance without explicitly modeling refracted light paths or estimating refractive indices.

\paragraph{Transparent surfaces and depth extraction.}
TSGS~\citep{li2025tsgs} separates geometry learning from transparency-aware appearance refinement using normal and de-lighting priors and a first-surface depth rule. TSPE-GS~\citep{xu2026tspe} and UniGS~\citep{min2026unigs} represent multiple depths per ray; G-GSTR~\citep{tian2026geometryaware} and TSTR~\citep{tian2026specular} add geometric or specular constraints. ClearForm focuses on how transparency changes the interface's geometric contribution during plane-based rendering and optimization.

\paragraph{Geometry--appearance factorization and scene decomposition.}
RT-Splatting~\citep{shi2026rt} factorizes geometric occupancy and optical opacity for surface-attribute aggregation in deferred reflection and volumetric transmission. GLINT~\citep{na2026glint} represents interface, transmission, and reflection using separate Gaussian components and secondary-ray transport. AlphaSurf and Car-GS also distinguish geometric from optical roles~\citep{wu2025alphasurf,li2025cargs}; related appearance methods include TR-Gaussians, RT-GS, and TransmissiveGS~\citep{liu2025tr,zeng2026rtgs,liang2026transmissive}. ClearForm investigates the geometric consequences of sharing transparency-dependent alpha, including the depth shift caused by overlapping interface and background planes. We derive the associated gradient and rasterization effects and quantify interface support using fixed-Gaussian comparisons. Our geometry-preserving dual-opacity formulation combines transparency-independent plane compositing, transmitted appearance, and sparse reflection on shared planar Gaussians with transparency-aware PGSR constraints.

\paragraph{Gaussian pruning.}
LightGaussian and PUP 3D-GS prune by significance or reconstruction sensitivity~\citep{fan2024lightgaussian,hanson2025pup}, while MaskGaussian preserves gradients for masked primitives~\citep{liu2025maskgaussian}. FastGS and birth--death structural learning use multi-view reconstruction signals~\citep{ren2026fastgs,tran2026birthdeath}. RT-Splatting~\citep{shi2026rt} prunes by geometric occupancy to preserve transparent surfaces. In dual-opacity reconstruction, a geometry-visible transparent interface can occlude background Gaussians in the geometry pass even though they contribute to transmitted RGB. Opacity-based pruning may discard such Gaussians when their learned opacity is low, while geometry-only visibility cannot capture their RGB usefulness. ClearForm's Dual-Evidence Lifecycle (DEL) instead retains primitives supported by geometry-pass visibility or positive first-order RGB-loss contribution across distinct cameras. Lineage aggregation and temporary evidence inheritance extend this criterion to clone/split children.

\begin{figure*}[t]
\centering
\includegraphics[width=\textwidth]{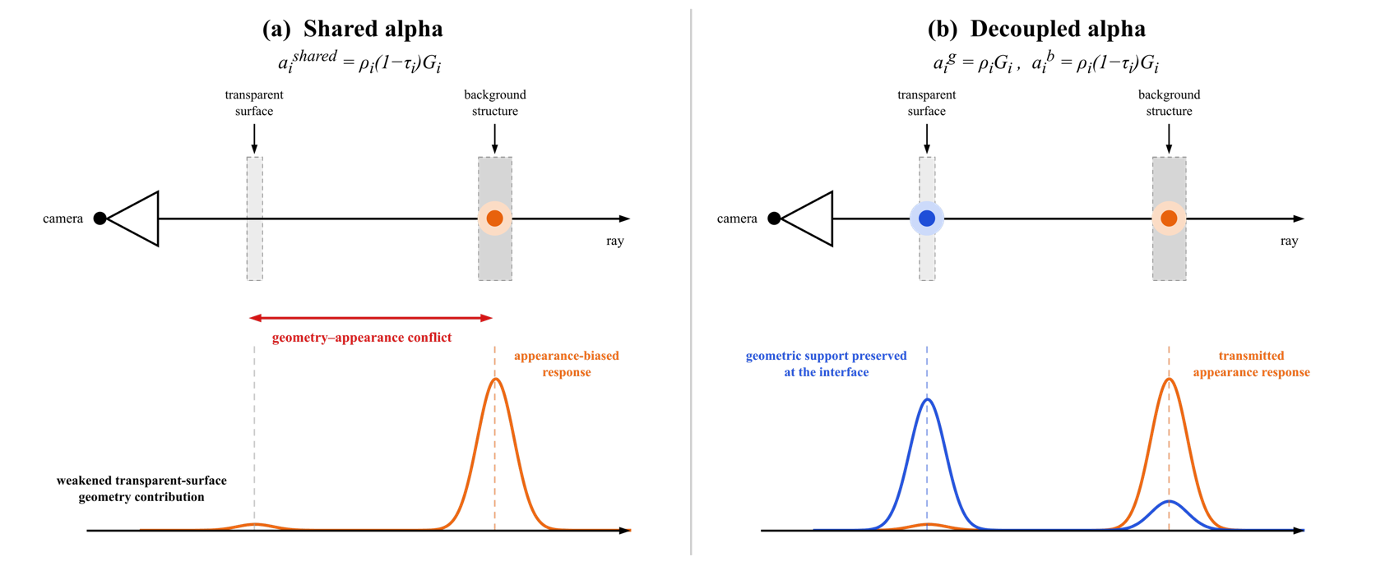}
\caption{\textbf{Geometry--Appearance Compositing Dilemma.} (a) Shared transparency-dependent alpha can weaken the interface in geometry compositing while revealing transmitted appearance. (b) Separate geometry and RGB alpha retain the interface in plane-attribute rendering while preserving transmitted appearance.}
\label{fig:gacd}
\end{figure*}

\section{Geometry--Appearance Compositing Dilemma}
\label{sec:analysis}
GACD arises when the same transparency-dependent alpha
is used for both appearance rendering and geometric
compositing. Reducing a transparent interface's alpha
reveals background appearance but also weakens its
contribution to geometric rendering. We analyze this
conflict in PGSR, which estimates depth from
alpha-blended plane normals and distances.
Specifically, we examine how shared alpha affects
blended-plane depth, a local geometric gradient
pathway, and the inclusion of interface contributions
during rasterization. \Cref{fig:gacd} illustrates
the corresponding interface and background contributions.

\subsection{Standard 3D Gaussian Splatting}
In 3DGS~\citep{kerbl2023}, Gaussian $i$ has world-space center $\mu_i\in\mathbb R^3$, rotation $R_i$, positive axis scales $ s_i$, opacity $\rho_i\in(0,1)$, and view-dependent color $ c_i$ evaluated for the current viewing direction. Its world-space covariance is $\Sigma_i=R_i\operatorname{diag}( s_i^2)R_i^\top$. Let $Q$ and $ t$ be the world-to-camera rotation and translation, and $J_i$ the $2\times3$ Jacobian of perspective projection at the Gaussian center. The projected covariance is
\begin{equation}
\Sigma_{i,\mathrm{2D}}=J_iQ\Sigma_iQ^\top J_i^\top.
\end{equation}
Let $ u_i$ be the projected center. The screen-space footprint $G_i(p)$ measures the influence of Gaussian $i$ at pixel $p$ before opacity:
\begin{equation}
G_i(p)=\exp\!\left[-\tfrac12(p- u_i)^\top
\Sigma_{i,\mathrm{2D}}^{-1}(p- u_i)\right].
\label{eq:footprint}
\end{equation}
With Gaussians sorted front to back by camera-space depth, alpha compositing uses
\begin{equation}
\begin{aligned}
a_i(p)&=\rho_iG_i(p),\\
T_i(p)&=\prod_{j<i}[1-a_j(p)],\qquad w_i(p)=T_i(p)a_i(p),
\end{aligned}
\label{eq:original-alpha}
\end{equation}
to obtain the Gaussian color contributions
\begin{equation}
\widehat I(p)=\sum_iw_i(p) c_i.
\label{eq:original-rgb}
\end{equation}
Here $a_i$ is the pixel-dependent blending opacity, $T_i$ is the incoming transmittance, and $w_i$ is the Gaussian's contribution weight. PGSR also uses these weights to aggregate plane attributes.

Standard 3DGS also adapts its representation during training: density control clones or splits primitives with large screen-space position gradients, prunes primitives with low learned opacity, and periodically resets opacity~\citep{kerbl2023}. This opacity-based pruning does not separately evaluate usefulness for geometry or RGB reconstruction. A Gaussian can contribute to either task despite low opacity. Alpha blending determines how a retained Gaussian contributes to the rendered result, whereas density control determines whether it remains in the representation.

\subsection{From RGB weights to PGSR surface geometry}
PGSR~\citep{chen2024pgsr} encourages local planarity by penalizing the smallest Gaussian axis scale: a flattened 3D ellipsoid approximates a surface patch, and the corresponding shortest-axis direction defines its normal. Let $\mu_i^{\rm cam}=Q\mu_i+ t$ be the camera-space center and $ n_i$ the oriented unit normal in that frame. The signed plane offset $d_i= n_i^\top\mu_i^{\rm cam}$ defines the plane $ n_i^\top x=d_i$. PGSR uses the alpha-compositing weights in \cref{eq:original-alpha} to blend plane normals and offsets at each pixel:
\begin{equation}
 N(p)=\sum_iw_i(p) n_i,\qquad
D(p)=\sum_iw_i(p)d_i.
\label{eq:original-planes}
\end{equation}
Following PGSR, the ray intersects the plane specified by these blended attributes at
\begin{equation}
Z(p)=\frac{D(p)}{ N(p)^\top r(p)},\qquad
 r(p)=K^{-1}\widetilde p.
\label{eq:plane-depth}
\end{equation}
Here $K$ denotes the current camera intrinsics, $\widetilde p=(p_x,p_y,1)^\top$ is the homogeneous pixel, and $ r$ is the unnormalized camera-space ray. Thus $Z$ is the optical-axis depth of the ray--plane intersection, not the alpha-weighted average of Gaussian center depths. The accumulated normal $ N$ is not unit-normalized; the ratio removes any common scale in $( N,D)$. Plane normals are oriented toward the camera; the renderer uses an equivalent sign-adjusted plane-distance convention. PGSR regularizes these geometric quantities across pixels and views.

\subsection{Geometric effects of shared alpha}
\paragraph{Shared-alpha formulation.}
Consider a transparency-aware extension of PGSR
in which material transparency $\tau_i\in[0,1]$
attenuates the same alpha used for both RGB
and geometric compositing:
\begin{equation}
a_i^{\rm shared}(p)=\rho_i(1-\tau_i)G_i(p).
\label{eq:shared-alpha}
\end{equation}
Increasing $\tau_i$ reduces the interface's
contribution to RGB compositing, allowing background
color to pass through. However, because the same
alpha also determines geometric compositing weights,
it reduces the interface's contribution to geometry
and changes the transmittance to background planes.
When multiple planes contribute to a pixel, these
changes can shift the estimated depth.

\paragraph{Gradient through projected Gaussian weights.}
Let $L_{\rm geo}$ be a differentiable loss on rendered geometry, and let $\theta_i$ denote the center, rotation, or scale parameters. Its gradient decomposes as
\begin{equation}
\begin{aligned}
\nabla_{\theta_i}L_{\rm geo}
={}&\sum_p\frac{\partial L_{\rm geo}}{\partial a_i(p)}
\underbrace{\rho_i(1-\tau_i)}_{\partial a_i/\partial G_i}
\nabla_{\theta_i}G_i(p)+{\mathcal R}_i.
\end{aligned}
\label{eq:gradient}
\end{equation}
The loss derivative $\partial L_{\rm geo}/\partial a_i$ includes effects on later transmittance and the rendered geometric attributes. The remainder ${\mathcal R}_i$ contains other geometric derivatives, including changes to plane distance and normal. Only the local pathway through the projected Gaussian weight $G_i$ has the explicit factor $1-\tau_i$. Compensating by increasing $\rho_i$ also increases base RGB opacity, coupling geometric support to optical attenuation.

\paragraph{Sensitivity of plane depth to alpha weights.}
For a single valid plane, alpha cancels in \cref{eq:plane-depth}, giving $Z=d_i/( n_i^\top r)$. When interface and background planes contribute to one pixel, changing their relative weights shifts the blended plane and hence the estimated depth. Thus alpha compositing can change multi-plane depth even when single-plane depth is invariant; \cref{eq:gradient} concerns a local derivative path rather than the complete update. Appendix~A gives the corresponding depth derivative.

\paragraph{Per-pixel alpha thresholding.}
Rasterization skips Gaussian contributions whose
per-pixel alpha falls below a threshold
$\alpha_{\min}$. At a fixed footprint, an interface
satisfying
\begin{equation}
\rho_iG_i(p)\geq\alpha_{\min}>\rho_i(1-\tau_i)G_i(p)
\label{eq:cutoff}
\end{equation}
can contribute to geometry blending with unattenuated alpha but is rejected under the shared-alpha rule in \cref{eq:shared-alpha}. The left-hand side is the unattenuated Gaussian alpha and the right-hand side is its transparency-attenuated counterpart. Passing the cutoff does not guarantee visibility after preceding Gaussians occlude the ray.

The three mechanisms play different roles: competing planes change the depth estimate, \cref{eq:gradient} identifies a locally transparency-scaled geometric derivative, and \cref{eq:cutoff} can remove interface contributions altogether. Thus a surface that transmits RGB can receive reduced weight in geometry rendering. The fixed-state measurements in \cref{sec:gacdexp} quantify the compositing and threshold effects, without isolating the contribution of the gradient pathway to training.

\begin{figure*}[t]
\centering
\includegraphics[width=\textwidth]{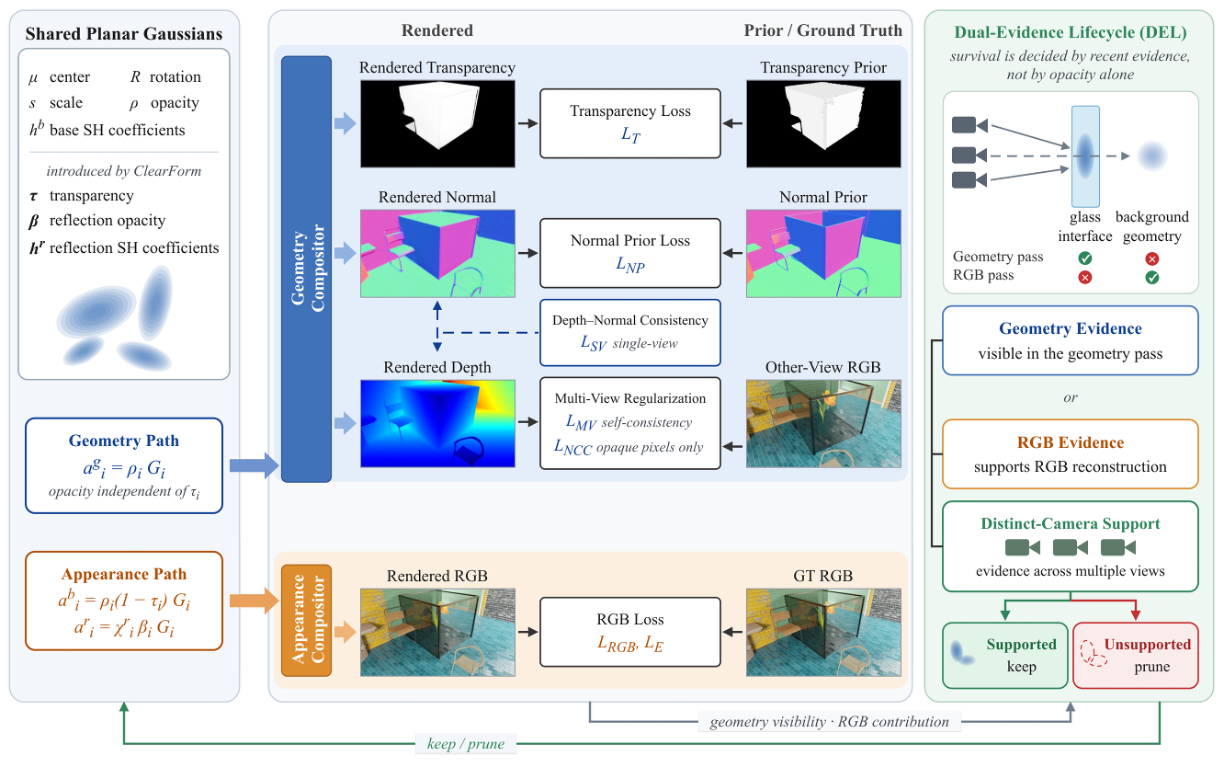}
\caption{\textbf{Overview of ClearForm.} Shared planar Gaussians feed separate geometry and appearance compositors, preserving the transparent interface in geometric rendering while reproducing transmitted color and reflection. Transparency-aware PGSR constraints and priors supervise geometry; image losses supervise appearance. Dual-Evidence Lifecycle (DEL) uses visibility in geometric rendering and estimated RGB-loss changes across distinct recently sampled camera views to guide primitive survival.}
\label{fig:overview}
\end{figure*}
\section{ClearForm}
\label{sec:method}
To prevent transmission from suppressing interface plane contributions as analyzed in \cref{sec:analysis}, ClearForm separates geometry and appearance compositing over shared planar Gaussians (\cref{sec:shared}). Transparency-aware PGSR constraints supervise the interface geometry (\cref{sec:surface}), while RGB rendering preserves transmitted color and sparse reflection. DEL addresses the complementary survival requirements of interface and background Gaussians (\cref{sec:del}). \Cref{fig:overview} summarizes the pipeline.

\subsection{Shared geometry and dual-opacity compositing}
\label{sec:shared}
Our inputs are calibrated multi-view RGB images, per-view surface normal priors, and soft transparency masks. We reconstruct transparent-interface geometry and view-dependent appearance alongside opaque scene geometry. Gaussian $i$ carries $(\mu_i,R_i, s_i,\rho_i,\tau_i, h_i^b)$, where $ h_i^b$ are base-color spherical-harmonic (SH) coefficients. Selected Gaussians additionally carry reflection opacity $\beta_i\in(0,1)$ and independent SH coefficients $ h_i^r$. The base and reflection components share center, rotation, and scale. The shared coefficient $\rho_i$ is a splatting opacity, not physical density; $\tau_i$ attenuates only base RGB alpha. Superscripts $g$, $b$, $r$, and $c$ denote geometry, base color, reflection, and combined color.

\paragraph{Geometry compositor.}
Both paths use the front-to-back alpha blending in \cref{eq:original-alpha}; they differ in per-pixel alpha and the attributes being blended. The geometry pass retains PGSR's alpha for plane attributes, whereas the RGB pass attenuates it according to transparency. The geometry-pass alpha and contribution weights are
\begin{equation}
\begin{aligned}
a_i^g(p)&=\rho_iG_i(p),\\
w_i^g(p)&=a_i^g(p)\prod_{j<i}[1-a_j^g(p)].
\end{aligned}
\end{equation}
Front-to-back blending yields accumulated geometric opacity $A^g$, rendered transparency $\widehat\tau$, and plane attributes $ N,D$:
\begin{align}
A^g&=\sum_iw_i^g, & \widehat\tau&=\sum_iw_i^g\tau_i,\\
 N&=\sum_iw_i^g n_i, & D&=\sum_iw_i^gd_i.
\label{eq:georeadout}
\end{align}
Depth is obtained with \cref{eq:plane-depth}. The rendered transparency $\widehat\tau$ is a coverage-weighted sum, not normalized by $A^g$. Its prior therefore constrains $\tau_i$ together with geometric opacity and footprint support. At fixed geometry and opacity, this geometry compositor does not attenuate interface normals or plane offsets when $\tau_i$ increases.

\paragraph{Appearance rendering and sparse reflection.}
\label{sec:appearance}
The base appearance alpha is
\begin{equation}
a_i^b(p)=\rho_i(1-\tau_i)G_i(p).
\end{equation}
$\chi_i^r\in\{0,1\}$ indicates whether Gaussian $i$'s reflection component is active at the current iteration. The corresponding reflection alpha is
\begin{equation}
a_i^r(p)=\chi_i^r\beta_iG_i(p).
\end{equation}
Independent SH coefficients define view-dependent base color $ c_i^b$ and reflection color $ c_i^r$, evaluated for the current viewing direction. At each Gaussian, base color is alpha-blended first and reflection uses the remaining transmittance:
\begin{align}
T_i^c &= \prod_{j<i}(1-a_j^b)(1-a_j^r),\\
\widehat I &=
\sum_i T_i^c\left[
a_i^b c_i^b
+(1-a_i^b)a_i^r c_i^r
\right].
\label{eq:rgb}
\end{align}
The effective alpha $a_i^{\rm eff}=1-(1-a_i^b)(1-a_i^r)$ accounts for base color and reflection without directly adding their alphas; reflection can also attenuate transmitted appearance. Reflection opacity is not multiplied by $1-\tau_i$, allowing reflection from highly transparent surfaces.

We add reflection parameters to Gaussians with sufficient transparency and geometric opacity unless their observed mean mask support indicates an opaque region. Once selected, a reflection component is retained, but its rendering activation $\chi_i^r$ additionally requires current transparency support; Appendix~\ref{ap:implementation} specifies these conditions.

\subsection{Transparency-aware PGSR regularization}
\label{sec:surface}

Building on PGSR's geometric regularization~\citep{chen2024pgsr}, we adapt surface supervision to transparent regions, where observed RGB texture may originate from background surfaces. Following PGSR~\citep{chen2024pgsr}, we encourage planar Gaussians by penalizing their smallest axis scale: $L_{\rm flat}=\operatorname{mean}_{i\in\mathcal V} \min_{\ell\in\{1,2,3\}}s_{i\ell}$, where $\mathcal V$ denotes the visible Gaussians.

\paragraph{Normal prior.}
We supervise the rendered normal $ N$ with a
predicted normal prior $ n^\star$.
Following the prior-rejection strategy used in
TSGS~\citep{li2025tsgs}, we exclude inconsistent
priors after an initial warm-up.
Let $V_{\rm NP}(p)$ indicate valid prior pixels,
additionally requiring
$ N(p)^\top n^\star(p)>\theta_{\rm NP}$
after warm-up. With the PGSR-style image-edge weight
$\omega_O(p)=(1-g_{\rm RGB}(p))^2$, we use
\begin{equation}
L_{\rm NP}=
\frac{\sum_p V_{\rm NP}(p)\omega_O(p)
\norm{ N(p)- n^\star(p)}_1}
{\sum_p V_{\rm NP}(p)+\epsilon_{\rm NP}}.
\label{eq:np}
\end{equation}
Here $g_{\rm RGB}$ is the normalized RGB gradient
magnitude, and $\epsilon_{\rm NP}>0$ stabilizes
the denominator. Unlike a cosine-based normal loss,
this formulation compares the unnormalized,
alpha-composited normal $ N$ with the prior;
the acceptance test likewise uses their raw
inner product.

\paragraph{Single-view depth--normal consistency.}
PGSR encourages agreement between rendered normals
and normals derived from neighboring depth samples,
using image gradients to reduce regularization near
edges. However, RGB edges in transparent regions
may come from background texture rather than the
interface. We therefore replace the RGB edge weight
with a transparency-mask edge weight in these regions.

Let $m$ be the transparency prior,
$g_m$ its normalized gradient magnitude, and
$M(p)=\ind[m(p)>\theta_m]$, with $\theta_m=0.5$.
The resulting weights are
\begin{align}
\omega_T&=(1-g_m)^2,\\
\omega&=(1-M)\omega_O+M\omega_T.
\label{eq:svweight}
\end{align}
Using the depth-derived normal $ n_D$, we define
\begin{equation}
L_{\rm SV}=\operatorname{mean}_p
\omega(p)
\norm{\sg(A^g(p)) n_D(p)- N(p)}_1.
\label{eq:svloss}
\end{equation}
The detached geometric coverage $\sg(A^g)$
matches the scale of the unnormalized rendered
normal $ N$. This adaptation reduces the
influence of transmitted texture on interface
geometry regularization.

\paragraph{Multi-view regularization.}
Following PGSR~\citep{chen2024pgsr}, we compute
multi-view geometric consistency using round-trip
reprojection error $e(p)$:

\begin{equation}
L_{\rm MV}=\operatorname{mean}_{p\in\mathcal C}
\sg(\exp[-e(p)])e(p).
\label{eq:mv}
\end{equation}
Here $\mathcal C$ excludes invalid depth and
reprojection errors above $\theta_{\rm rep}$,
and $\sg$ denotes stop-gradient. Transparent
pixels remain eligible for geometric consistency.

For photometric consistency, we use PGSR's
plane-induced homography and normalized
cross-correlation loss $L_{\rm NCC}$, but restrict
matching to low-transparency regions:
\begin{equation}
\mathcal C_{\rm NCC}=
\{p\in\mathcal C:
\widehat\tau(p)<\theta_T
\ \land\ m(p)\leq\theta_m\}.
\end{equation}
We use $\theta_T=0.5$. This restriction avoids
matching background texture as if it belonged to
the transparent interface.

\paragraph{Image and material constraints.}
We use the standard RGB reconstruction objective
\begin{equation}
\begin{aligned}
L_{\rm RGB}={}&
(1-\lambda_{\rm SSIM})\norm{\widehat I-I}_1\\
&+\lambda_{\rm SSIM}
[1-\operatorname{SSIM}(\widehat I,I)].
\end{aligned}
\end{equation}
The image loss covers all pixels, including
transmitted background appearance.

We supervise rendered transparency $\widehat\tau$
using the scaled mask target
$m^\star=\clip(\kappa_m m,0,1)$:
\begin{equation}
L_T=u_T(k)
[1-\operatorname{SSIM}(\widehat\tau,m^\star)].
\label{eq:transprior}
\end{equation}
Here $u_T(k)=\min(1,k/T_{\rm warm})$ gradually
activates the transparency prior.
The original mask $m$, rather than $m^\star$,
is used for geometric weighting and NCC masking.
Gradients from $L_T$ to Gaussian geometry and
densification are separately scaled.

We additionally apply a boundary-weighted RGB
gradient loss $L_E$ to improve appearance
consistency near transparency boundaries.
Its formulation and optimization schedule
are provided in Appendix~\ref{ap:implementation}.

\begin{figure*}[t]
\centering
\includegraphics[width=\linewidth]{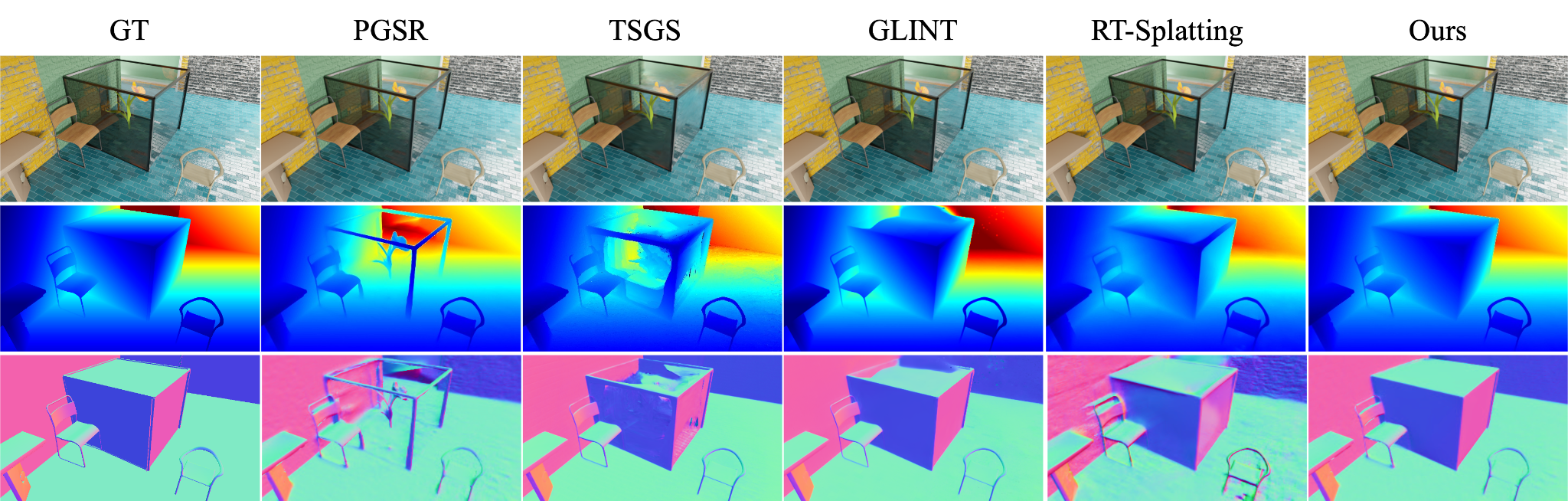}
\caption{\textbf{3D-FRONT-T comparison.} ClearForm recovers the glass interface in depth and normals while preserving the transmitted scene appearance. Columns show ground truth, PGSR, TSGS, GLINT, RT-Splatting, and ClearForm; rows show RGB, depth, and normals.}
\label{fig:frontqualfinal}
\end{figure*}

% BEGIN tables/front.tex
\begin{table*}[t]
\centering
\caption{\textbf{3D-FRONT-T reconstruction and novel-view synthesis.} Five-scene means. Unstarred baseline scores are quoted from GLINT~\citep{na2026glint}; $^*$ denotes a local reproduction.}
\label{tab:front}
\small\setlength{\tabcolsep}{3.5pt}
\resizebox{\textwidth}{!}{%
\begin{tabular}{lrrrrrrrrrrr}
\toprule
\multirow{2}{*}{Method} & \multicolumn{3}{c}{Normal} & \multicolumn{3}{c}{Depth} & \multicolumn{2}{c}{Mesh} & \multicolumn{3}{c}{Novel View}\\
\cmidrule(lr){2-4}\cmidrule(lr){5-7}\cmidrule(lr){8-9}\cmidrule(lr){10-12}
& MAE$\downarrow$ & $11.25^\circ\uparrow$ & $22.5^\circ\uparrow$ & AbsRel$\downarrow$ & RMSE$\downarrow$ & $\delta<1.25\uparrow$ & CD (dm)$\downarrow$ & F1$\uparrow$ & PSNR$\uparrow$ & SSIM$\uparrow$ & LPIPS$\downarrow$\\
\midrule
2DGS~\citep{huang2024} & 25.97 & 52.19 & 64.55 & 0.200 & 0.240 & 76.71 & 0.85 & 0.688 & 32.26 & 0.940 & 0.080\\
PGSR$^*$~\citep{chen2024pgsr} & 26.01 & 57.17 & 66.23 & 0.198 & 0.251 & 77.18 & 0.59 & 0.786 & 32.25 & 0.949 & \rankthree{0.065}\\
Ref-GS~\citep{zhang2025refgs} & 41.55 & 18.61 & 36.79 & 0.280 & 0.360 & 57.52 & 1.29 & 0.408 & 32.30 & 0.940 & 0.090\\
EnvGS~\citep{xie2025envgs} & 14.37 & 68.22 & 80.23 & 0.130 & 0.160 & 86.10 & 0.87 & 0.640 & 33.71 & 0.940 & 0.070\\
TSGS~\citep{li2025tsgs} & \rankthree{9.89} & \rankthree{86.29} & \rankthree{92.24} & 0.080 & 0.120 & 95.56 & \rankthree{0.52} & \rankthree{0.798} & 28.80 & 0.870 & 0.140\\
GLINT~\citep{na2026glint} & \ranktwo{7.96} & \ranktwo{86.37} & \rankone{92.28} & \ranktwo{0.040} & \rankone{0.070} & \ranktwo{98.32} & \ranktwo{0.34} & \ranktwo{0.836} & \ranktwo{34.50} & \rankone{0.960} & \rankone{0.050}\\
RT-Splatting$^*$~\citep{shi2026rt} & 18.43 & 59.00 & 76.04 & \rankthree{0.046} & \rankthree{0.084} & \rankthree{97.98} & 0.57 & 0.691 & \rankone{35.23} & \ranktwo{0.959} & \rankone{0.050}\\
\midrule
\method{} & \rankone{7.41} & \rankone{87.28} & \ranktwo{92.27} & \rankone{0.027} & \ranktwo{0.074} & \rankone{98.40} & \rankone{0.27} & \rankone{0.928} & \rankthree{34.14} & \rankthree{0.956} & \ranktwo{0.054}\\
\bottomrule
\end{tabular}}
\end{table*}

% END tables/front.tex

\subsection{Dual-Evidence Lifecycle}
\label{sec:del}
Dual-opacity compositing creates different contribution weights for geometry and RGB rendering. A transparent interface can contribute strongly to geometry compositing, reducing the accumulated transmittance to background Gaussians. Consequently, these Gaussians may receive small geometry-pass weights while remaining useful for transmitted RGB, where interface alpha is attenuated by transparency. Geometry-based visibility criteria may therefore overlook their appearance contribution, while conventional opacity-based pruning can remove them when their learned opacity is low. We introduce Dual-Evidence Lifecycle (DEL), which determines Gaussian survival using geometry-pass visibility or positive RGB-loss contribution across distinct cameras.

\paragraph{Geometry-pass visibility.}
We identify Gaussians sufficiently visible under the current geometry-pass alpha. Let $T_i^g(p)=\prod_{j<i}[1-a_j^g(p)]$ be the incoming geometric transmittance at pixel $p$, distinct from the color transmittance $T_i^c$ in \cref{eq:rgb}. Gaussian $i$ is geometry-visible in camera $v$ if some pixel has both a non-negligible geometric alpha $a_i^g(p)$ and sufficient incoming transmittance $T_i^g(p)$. The first condition requires a supported footprint; the second excludes contributions substantially occluded by earlier Gaussians. We record the camera whenever such an observation occurs. This is a visibility criterion under the current reconstruction, not a test of surface accuracy or multi-view depth consistency.

\paragraph{Signed first-order RGB contribution.}
Let $o_i^b=\rho_i(1-\tau_i)$ denote the base-color
opacity coefficient before footprint evaluation.
For camera $v$, we estimate Gaussian $i$'s
contribution to RGB reconstruction as

\begin{equation}
\Delta_i^v=-o_i^b\frac{\partial L_{\rm RGB}^v}{\partial o_i^b}
-\beta_i\frac{\partial L_{\rm RGB}^v}{\partial\beta_i}.
\label{eq:score}
\end{equation}
The signed first-order estimate $\Delta_i^v$ in \cref{eq:score} approximates the RGB loss change when the Gaussian's base and reflection opacity coefficients are reduced to zero. A positive value predicts a loss increase, while a negative value predicts a decrease. The estimate holds the other parameters fixed but includes downstream compositing effects in its derivatives. Inactive reflection contributes zero. We evaluate only the RGB reconstruction loss, excluding geometry, material-prior, and boundary penalties.

Let $\mathcal F$ denote a clone/split lineage and $\mathcal O_v^{\rm RGB}$ the Gaussians eligible for RGB evidence in camera $v$. Eligibility is determined using the incoming base-color transmittance $T_i^b(p)=\prod_{j<i}(1-a_j^b(p))$, rather than the combined color transmittance $T_i^c$; the RGB-loss score itself includes both base color and reflection. We aggregate signed loss-change estimates over the eligible lineage members:
\begin{equation}
\Delta_{\mathcal F}^v=\sum_{i\in\mathcal F\cap\mathcal O_v^{\rm RGB}}\Delta_i^v.
\label{eq:family}
\end{equation}
An observed Gaussian $i\in\mathcal O_v^{\rm RGB}$ receives RGB evidence when the lineage sum $\Delta_{\mathcal F}^v$ is positive. Negative contributions are retained in the sum, so the test reflects the lineage's net predicted effect on RGB reconstruction.

\paragraph{Recent distinct-camera support.}
For each Gaussian, we track camera-specific geometry and RGB evidence over a recent window of $W$ iterations. Let $\mathcal E_i^g$ and $\mathcal E_i^{\rm RGB}$ contain the distinct cameras whose latest recorded observation within that window supports the Gaussian in each pass. A later unsupported observation from the same camera revokes its earlier evidence. After an initial grace period, Gaussian $i$ survives a pruning step if
\begin{equation}
\left|\mathcal E_i^g\cup\mathcal E_i^{\rm RGB}\right|\geq K_{\min}.
\label{eq:survival}
\end{equation}
Here $K_{\min}$ is the distinct-camera threshold; a camera observed by both paths counts once. This union allows an interface to be supported by geometry observations and an occluded appearance-bearing primitive to be supported by RGB, while requiring evidence across views rather than from a single image. New Gaussians receive an initial grace period, and an observation-condition change permits a short deferral while additional cameras are sampled. Clone/split children receive provisional inherited evidence that expires unless replaced by their own observations; repeated splitting cannot extend this support indefinitely.

Opacity-threshold culling is disabled while DEL is active: pruning follows the distinct-camera evidence criterion, with temporary protection for newly created Gaussians. Opacity-reset protection is a separate, stricter rule for supported transparent Gaussians (Appendix~\ref{ap:implementation}).

\begin{figure*}[t]
\centering
\includegraphics[width=\linewidth]{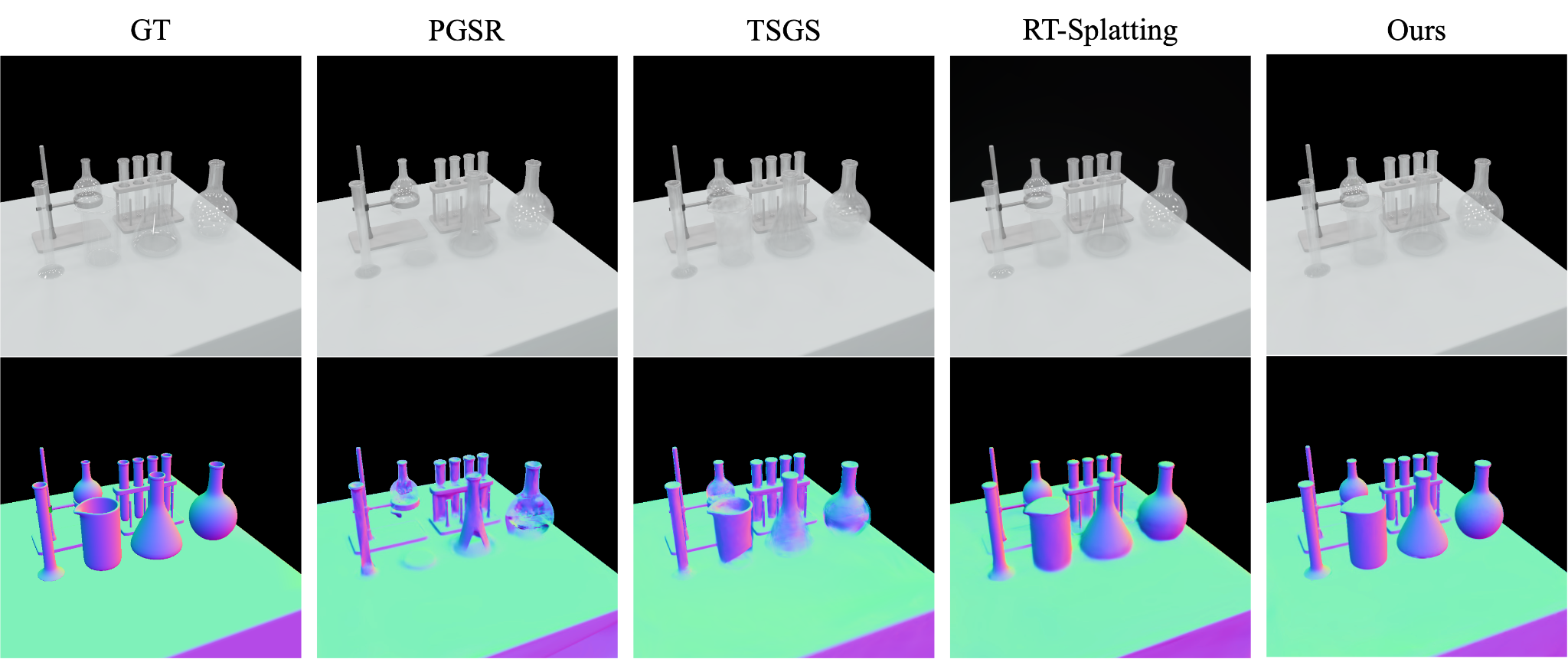}
\caption{\textbf{TransLab comparison.} ClearForm preserves transparent appearance while reconstructing coherent vessel surface normals. Columns show ground truth, PGSR, TSGS, RT-Splatting, and ClearForm; rows show RGB and normals.}
\label{fig:labqualfinal}

\end{figure*}

\subsection{Training objective and optimization}
We optimize the weighted objective
\begin{equation}
\begin{aligned}
L={}&L_{\rm RGB}+\lambda_{\rm flat}L_{\rm flat}
+\lambda_{\rm NP}L_{\rm NP}+\lambda_{\rm SV}L_{\rm SV}\\
&+\lambda_{\rm MV}L_{\rm MV}+\lambda_{\rm NCC}L_{\rm NCC}
+\lambda_TL_T\\
&+\lambda_\tau L_\tau+\lambda_E L_E,
\end{aligned}
\label{eq:objective}
\end{equation}
The coefficients $\lambda$ balance surface and material constraints, while $L_{\rm RGB}$ uses the image-term mixture defined above.

Both datasets are trained for 30,000 iterations at half input resolution. Normal guidance starts immediately, with the inner-product acceptance test added after iteration 3,000. Single-view geometry, multi-view geometry, and NCC begin after iteration 7,000. Reflection candidates are considered every 100 iterations from iteration 7,000 to 15,000. Densification and evidence-based pruning end before iteration 15,000, followed by continuous optimization of the retained primitives. Appendix~\ref{ap:implementation} reports additional optimization settings and DEL implementation details.

\section{Experiments}
\label{sec:experiments}

\subsection{Experimental setup}
\paragraph{Datasets and implementation.} We evaluate five 3D-FRONT-T indoor scenes~\citep{na2026glint} and eight TransLab glassware scenes~\citep{li2025tsgs}. The former contains scene-scale transparent interfaces and opaque contents with ground-truth (GT) RGB and geometry; the latter is a Blender-rendered benchmark of laboratory glassware with thin and curved surfaces. Our implementation builds on the official PGSR codebase~\citep{pgsrcode}. On both datasets, PGSR$^*$ denotes our reproduction using the unmodified official implementation. We precompute normal priors using StableNormal~\citep{ye2024stablenormal} and transparency masks using SAM~3~\citep{carion2026sam3} with the fixed text prompt ``transparent glass''. These priors remain fixed during optimization and are used for both benchmark and real-scene experiments. GT geometry and transparent-region annotations are reserved for evaluation. On TransLab, the supplied foreground alpha is used to mask normal supervision and composite GT RGB onto a black background. It is not used as a transparency prior, and RGB training includes all pixels. For qualitative evaluation, we use Ref-DL3DV, the reflective and transparent DL3DV-10K subset used by GLINT~\citep{na2026glint,ling2024dl3dv}. These scenes have no GT geometry. Additional held-out views are presented in Appendix~\ref{ap:real}.

\paragraph{Evaluation.} We follow the benchmark evaluation protocols of GLINT~\citep{na2026glint} for 3D-FRONT-T and TSGS~\citep{li2025tsgs} for TransLab. On 3D-FRONT-T, we evaluate surface normals using mean angular error (MAE) and angular accuracy at $11.25^\circ$ and $22.5^\circ$, depth using AbsRel, normalized RMSE, and $\delta<1.25$, and mesh reconstruction using Chamfer distance (CD) and F1. Depth evaluation follows GLINT's per-view median scale alignment. On TransLab, we evaluate mesh reconstruction using CD and F1, following TSGS. The F1 distance thresholds are 5\,cm for 3D-FRONT-T and 5\,mm for TransLab. For both datasets, novel-view synthesis is evaluated using PSNR, SSIM, and LPIPS~\citep{wang2004ssim,zhang2018lpips}. Published baseline results are taken from GLINT on 3D-FRONT-T and TSGS on TransLab. The $^*$ symbol denotes our local reproductions using official implementations. Our meshes are extracted from geometry-pass depth using PGSR's TSDF fusion~\citep{pgsrcode}, with voxel sizes of 0.01\,m and 0.002\,m for 3D-FRONT-T and TransLab, respectively. While the benchmark evaluation protocols are shared, methods may differ in their training priors and reconstruction pipelines.

\begin{table}[!ht]
\centering
\caption{\textbf{TransLab reconstruction and novel-view synthesis.} Eight-scene means. Published baseline scores are quoted from TSGS~\citep{li2025tsgs}; $^*$ denotes a local reproduction. PGSR$^*$ RGB uses the common foreground-masked full-frame evaluation.}
\label{tab:lab}
\footnotesize\setlength{\tabcolsep}{3pt}
\begin{tabular}{lrrrrr}
\toprule
\multirow{2}{*}{Method} & \multicolumn{2}{c}{Mesh} & \multicolumn{3}{c}{Novel View}\\
\cmidrule(lr){2-3}\cmidrule(lr){4-6}
& CD (mm)$\downarrow$ & F1$\uparrow$ & PSNR$\uparrow$ & SSIM$\uparrow$ & LPIPS$\downarrow$\\
\midrule
SuGaR~\citep{guedon2024} & 17.27 & 0.440 & 23.03 & 0.941 & 0.123\\
2DGS~\citep{huang2024} & 3.55 & 0.840 & 36.91 & 0.924 & 0.167\\
GOF~\citep{yu2024gof} & 8.04 & 0.660 & 24.06 & 0.931 & 0.131\\
PGSR$^*$~\citep{chen2024pgsr} & 7.11 & 0.867 & 36.66 & 0.988 & \rankthree{0.017}\\
TSGS~\citep{li2025tsgs} & \ranktwo{1.85} & \ranktwo{0.950} & \ranktwo{39.08} & \rankthree{0.989} & 0.020\\
RT-Splatting$^*$~\citep{shi2026rt} & \rankthree{2.64} & \rankthree{0.900} & \rankthree{38.28} & \ranktwo{0.990} & \rankone{0.014}\\
\midrule
\method{} & \rankone{1.65} & \rankone{0.963} & \rankone{40.14} & \rankone{0.991} & \rankone{0.014}\\
\bottomrule
\end{tabular}
\end{table}

% BEGIN figures/qualitative_ablation.tex
\begin{figure*}[t]
\centering
\includegraphics[width=\textwidth]{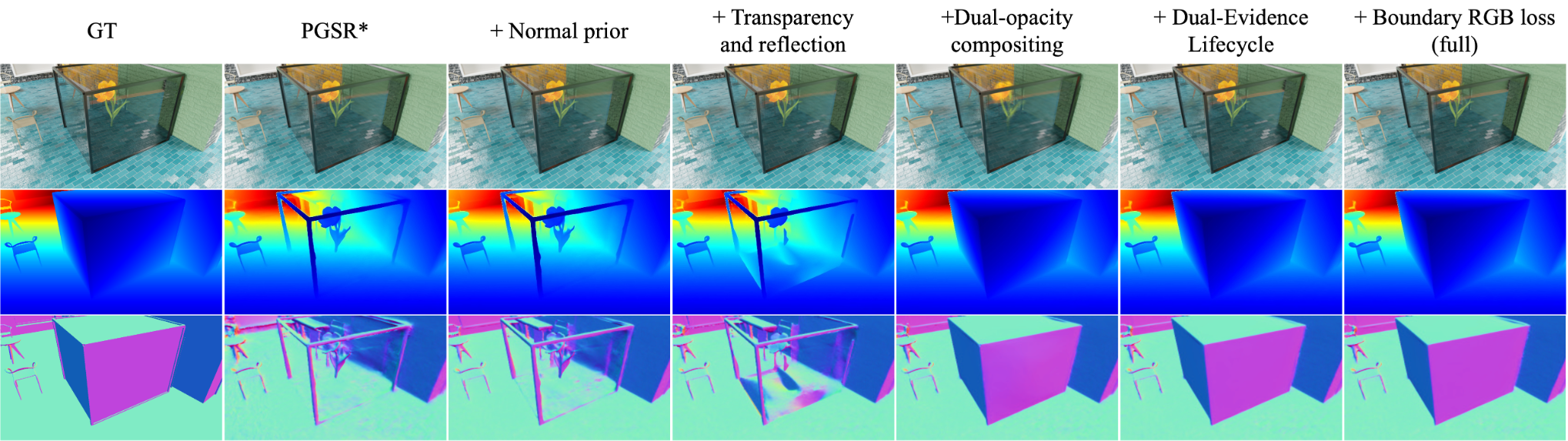}
\caption{\textbf{Cumulative ablation on 3D-FRONT-T.} Dual-opacity compositing restores the transparent interface left distorted by transparency and reflection modeling under shared opacity. From left to right, the columns progress from ground truth and PGSR to the full model; rows show RGB, depth, and normals.}
\label{fig:qualablation}
\smallskip
\captionof{table}{\textbf{Cumulative ablation on 3D-FRONT-T.} Each row adds one component to the preceding configuration. MAE is in degrees, accuracies in percent, and CD in mm.}
\label{tab:ablation}
\footnotesize\setlength{\tabcolsep}{2.3pt}
\begin{tabular*}{\textwidth}{@{\extracolsep{\fill}}lrrrrrrrrrrr@{}}
\toprule
\multirow{2}{*}{Configuration} & \multicolumn{3}{c}{Normal} & \multicolumn{3}{c}{Depth} & \multicolumn{2}{c}{Mesh} & \multicolumn{3}{c}{Novel View}\\
\cmidrule(lr){2-4}\cmidrule(lr){5-7}\cmidrule(lr){8-9}\cmidrule(lr){10-12}
& MAE$\downarrow$ & $11.25^\circ\uparrow$ & $22.5^\circ\uparrow$ & AbsRel$\downarrow$ & RMSE$\downarrow$ & $\delta<1.25\uparrow$ & CD (mm)$\downarrow$ & F1$\uparrow$ & PSNR$\uparrow$ & SSIM$\uparrow$ & LPIPS$\downarrow$\\
\midrule
PGSR$^*$ & 26.01 & 57.17 & 66.23 & 0.198 & 0.251 & 77.18 & 59.18 & 0.786 & 32.25 & 0.949 & 0.065\\
+ Normal prior & 21.84 & 65.13 & 72.44 & 0.195 & 0.249 & 78.05 & 59.24 & 0.794 & 32.22 & 0.949 & 0.066\\
+ Transparency and reflection & 21.20 & 66.24 & 74.02 & 0.216 & 0.272 & 72.36 & 59.25 & 0.794 & 33.05 & 0.951 & 0.063\\
+ Dual-opacity compositing & 7.43 & 87.43 & 92.24 & 0.026 & 0.074 & 98.33 & 27.09 & 0.925 & 32.05 & 0.941 & 0.079\\
+ Dual-Evidence Lifecycle & 7.42 & 87.40 & 92.30 & 0.027 & 0.074 & 98.33 & 27.15 & 0.925 & 34.11 & 0.957 & 0.053\\
+ Boundary RGB loss (full) & 7.41 & 87.28 & 92.27 & 0.027 & 0.074 & 98.40 & 27.05 & 0.928 & 34.14 & 0.956 & 0.054\\
\bottomrule
\end{tabular*}

\end{figure*}
% END figures/qualitative_ablation.tex

\subsection{Benchmark comparisons}

\paragraph{3D-FRONT-T.}
\method{} reduces CD from 59.18 to \FrontCD{}\,mm (0.59 to 0.27\,dm in \cref{tab:front}) and raises F1 from 0.786 to 0.928 relative to our PGSR reproduction. Compared with the reported GLINT results, CD decreases from 0.34 to 0.27\,dm and F1 rises from 0.836 to 0.928. Against our local RT-Splatting reproduction, CD decreases from 0.57 to 0.27\,dm and normal MAE from $18.43^\circ$ to $7.41^\circ$, although RT-Splatting achieves higher PSNR (35.23 vs. 34.14\,dB). \Cref{fig:frontqualfinal} shows the interface in a matched view.

\paragraph{TransLab.}
On TransLab, \method{} obtains 1.65\,mm CD, 0.963 F1, and 40.14\,dB PSNR (\cref{tab:lab}), compared with 1.85\,mm, 0.950, and 39.08\,dB for TSGS. Relative to our local RT-Splatting reproduction, CD decreases from 2.64 to 1.65\,mm while PSNR increases from 38.28 to 40.14\,dB. The RGB and normal maps in \cref{fig:labqualfinal} show transparent appearance alongside coherent vessel surfaces. PGSR$^*$ has a large CD outlier in scene 06; Appendix~C gives all scene-level comparisons.

% BEGIN figures/qualitative_main.tex
% Author-supplied comparison montages; keep method order aligned with the captions.

% END figures/qualitative_main.tex

\subsection{Ablation study}
\label{sec:ablation}

\Cref{tab:ablation,fig:qualablation} trace the cumulative changes from PGSR to the full method in geometry and rendering.

\paragraph{Normal guidance and shared-opacity appearance.}
Normal guidance lowers MAE from $26.01^\circ$ to $21.84^\circ$, but mesh CD remains approximately 59.2\,mm. Adding transparency and reflection improves PSNR to 33.05\,dB while CD remains 59.25\,mm and AbsRel increases to 0.2164. Improved image fitting alone therefore does not resolve interface geometry in this shared-opacity configuration.

\paragraph{Dual-opacity compositing.}
Separating geometry and base RGB opacity reduces AbsRel from 0.2164 to 0.0263, normal MAE from $21.20^\circ$ to $7.43^\circ$, and CD from 59.25 to 27.09\,mm. F1 increases from 0.794 to 0.925. This is the largest reduction in CD and normal MAE in the cumulative ablation, linking the geometry-specific alpha rule to substantially improved surface reconstruction under matched supervision. PSNR falls to 32.05\,dB, so geometry improvement alone does not guarantee better RGB. As this is a training ablation, it includes both the compositing rule and its effects on the learned representation. The matched view shows a more coherent transparent interface in depth and normals (\cref{fig:qualablation}).

\paragraph{Dual-Evidence Lifecycle.}
Adding DEL raises PSNR by 2.06\,dB and reduces LPIPS from 0.0787 to 0.0529, while CD changes only from 27.09 to 27.15\,mm and F1 remains approximately 0.925. This gain is consistent with DEL preserving RGB-useful Gaussians behind geometrically visible interfaces under dual-opacity rendering, while maintaining similar aggregate surface accuracy. The cumulative ablation evaluates the complete lifecycle; it does not directly identify individual pruning events or isolate the effects of its evidence and lineage components.

\paragraph{Boundary RGB loss.}
The boundary term yields small changes in PSNR (34.11 to 34.14\,dB) and CD (27.15 to 27.05\,mm). Appendix~\ref{ap:ablation} separately examines the transparent-region geometry terms by leaving one out at a time.

\subsection{Effect of shared alpha on interface support}
\label{sec:gacdexp}

The preceding ablation shows that training with
decoupled geometry compositing substantially improves
surface reconstruction. To examine the compositing
effect separately from changes in the learned
representation, we evaluate the shared- and
decoupled-alpha geometry rules on the same Gaussian checkpoint.

Using ground-truth first-surface depth, we
identify Gaussians whose planes intersect near the
interface and measure their geometric contribution
weights and pixel coverage. Decoupled geometry
compositing increases mean first-hit weight from
0.0179 to 0.0401 and coverage from 7.08\% to
14.70\%. Under shared alpha, 44.95\% of first-hit
pixel--Gaussian pairs active under decoupled alpha
fall below the per-pixel alpha threshold.
These results directly quantify the loss of interface
support caused by shared alpha at a fixed Gaussian
representation.

Appendix~\ref{ap:gacd} defines these measurements
and reports a supplementary cross-evaluation of
shared- and decoupled-trained checkpoints. This
comparison distinguishes immediate compositing-rule
effects from differences between trained
representations, without isolating individual
gradient mechanisms.

\section{Conclusion and Discussion}
\label{sec:conclusion}
\method{} analyzes and addresses the loss of transparent-interface geometry when the same transparency-dependent alpha controls appearance and geometric compositing. GACD explains why alpha affects mixed-plane depth despite cancelling for an isolated plane, identifies a local geometric gradient pathway and per-pixel threshold effects, and quantifies interface support with ground-truth first-hit measurements. Geometry-preserving dual-opacity reconstruction then retains plane contributions from transparent interfaces while representing transmitted appearance and sparse reflection on shared planar Gaussians under transparency-aware PGSR constraints.

Dual-opacity compositing also exposes different survival requirements: transparent interfaces can occlude background Gaussians in the geometry pass while allowing their RGB contributions to pass through. DEL addresses this mismatch using geometry-pass visibility or predicted positive RGB-loss contribution across distinct recent views. The matched ablation shows that dual-opacity compositing gives the largest geometric gain, while DEL gives the largest subsequent rendering gain. These findings establish complementary roles for the reconstruction and lifecycle components, not a causal attribution to any single gradient path. Strong reflections and missing transparency guidance can degrade interface reconstruction, and single-depth mesh fusion can spuriously cap open containers. Across 3D-FRONT-T and TransLab, ClearForm improves transparent-surface reconstruction while preserving transmitted appearance and view-dependent reflection.

\section*{Acknowledgments}
This research was supported by the PoliceLab 3.0
Program through the Korea Institute of Police
Technology (KIPoT), funded by the Korean National
Police Agency (KNPA) (No.~RS-2025-25441773).

% END paper.tex

\FloatBarrier

\clearpage
\appendix
\setcounter{section}{0}
\setcounter{equation}{0}
\setcounter{table}{0}
\setcounter{figure}{0}
\numberwithin{equation}{section}
\numberwithin{table}{section}
\numberwithin{figure}{section}
\renewcommand{\theequation}{\thesection\arabic{equation}}
\renewcommand{\thetable}{\thesection\arabic{table}}
\renewcommand{\thefigure}{\thesection\arabic{figure}}
\section*{Appendices}
\begingroup
\fontsize{9.75}{10.5}\selectfont

% BEGIN supplement_body.tex

\section{GACD: Compositing and Plane-Depth Sensitivity}
\label{ap:derivation}
We derive the local sensitivity terms underlying the GACD analysis of \cref{sec:analysis}.

\subsection{Gradient through projected Gaussian weights}
At a fixed camera and pixel, let $ F=\sum_jw_j f_j$ composite an attribute $ f_j$ using the 3DGS weights. If $ F_{>i}=\sum_{j>i}a_j f_j\prod_{i<\ell<j}(1-a_\ell)$ is the composite behind primitive $i$, differentiation with fixed attributes and ordering gives
\begin{equation}
\frac{\partial F}{\partial a_i}=T_i( f_i- F_{>i}).
\label{eq:suffix}
\end{equation}
The suffix term accounts for the effect of $a_i$ on contributions behind Gaussian $i$. Under shared alpha, $\partial a_i/\partial G_i=\rho_i(1-\tau_i)$, so the projected-weight gradient pathway is attenuated without implying that the entire geometry gradient decreases by the same factor.

\subsection{Alpha-weight sensitivity of PGSR plane depth}
For PGSR plane depth $Z=D/( N^\top r)$, let $D_{>i}$ and $ N_{>i}$ denote the suffix composites of plane distance and normal. Applying \cref{eq:suffix} and the quotient rule gives
\begin{equation}
\frac{\partial Z}{\partial a_i}
=\frac{T_i[(d_i-D_{>i})-Z( n_i- N_{>i})^\top r]}
{ N^\top r}.
\label{eq:ratiojac}
\end{equation}
For a single contributing plane, its alpha weight cancels in the depth ratio. With two parallel planes at depths $z_f$ and $z_b$, however, $Z=(w_fz_f+w_bz_b)/(w_f+w_b)$; reducing the interface weight moves the estimate toward the background depth.

\section{Implementation Details}
\label{ap:implementation}
\paragraph{Optimization.}
The weights in \cref{eq:objective} are $(\lambda_{\rm flat},\lambda_{\rm NP},\lambda_{\rm SV},\lambda_{\rm MV},\lambda_{\rm NCC})=(100,0.01,0.015,0.03,0.15)$ and $(\lambda_T,\lambda_\tau,\lambda_E,\lambda_{\rm SSIM})=(1,10^{-4},0.05,0.2)$. For $L_T$, geometry/opacity gradients are scaled by $0.01$ (TransLab: $0.1$) and densification-statistic gradients by $0.005$ (TransLab: $0.05$); $\tau_i$ gradients are unaffected. TransLab normal priors are reversed. After 3,000 iterations, the normal-acceptance threshold is $0.3$; mask/transparency thresholds are $0.5$, and reprojection tolerance is one pixel.

\paragraph{Material and boundary losses.}
With mean observed mask support $\bar m_i$, the low-support transparency penalty is
\begin{equation}
L_\tau=u_T(k)\operatorname{mean}_i
[\kappa_O\ind[\bar m_i<\theta_O]\tau_i],
\label{eq:material-penalty}
\end{equation}
where $(\theta_O,\kappa_O)=(0.1,10)$ and $u_T(k)=\min(1,k/T_{\rm warm})$, with $T_{\rm warm}=5{,}000$. The transparency target is scaled by $\kappa_m=1.5$.

Let $h$ denote normalized RGB gradient magnitude. The boundary loss uses a detached spatial weight $W_E$ informed by transparency-mask boundaries, rendered geometry, and depth--normal variation:
\begin{equation}
L_E=u_E(k)\frac{\sum_p W_E(p)
|h(\widehat I)(p)-h(I)(p)|}
{\sum_p W_E(p)+\epsilon},
\label{eq:edge-loss}
\end{equation}
where $\epsilon>0$ and $u_E(k)=\clip((k-3{,}500)/3{,}500,0,1)$. The loss updates Gaussian parameters, including SH coefficients, but does not contribute to the screen-space densification statistic.

\paragraph{Sparse reflection.}
Candidates require $\tau_i\geq0.1$, $\rho_i\geq0.005$, and no confirmed low-mask support ($<0.1$). Reflection opacity starts at $\beta_i=0.02$. Selected parameters persist, with activation checked during rendering.

\paragraph{Dual-Evidence Lifecycle.}
Geometry evidence requires $\min(0.99,a_i^g)\geq1/255$ and $T_i^g>0.5$; RGB evidence uses base-color $T_i^b>0.5$. Each pass stores at most two camera records. DEL prunes every 1,000 iterations using evidence from at least two distinct cameras within a 3,000-iteration window; opacity-threshold pruning is disabled. New Gaussians have 3,000-iteration probation, and inherited clone/split evidence expires without revalidation. Supported transparent Gaussians with positive RGB evidence, validated geometry, or provisional transparent lineage are protected from opacity reset. Otherwise, opacity is reset to at most $0.05$ and low-mask Gaussians reset $\tau_i$ to $0.001$; split children inherit $\tau_i$ only when the parent has $\tau_i>0.1$.

\endgroup
\section{Per-Scene Results and Evaluation}
\label{ap:results}
\Cref{tab:labgeoper,tab:labnvsper} report per-scene TransLab geometry and rendering results, with equal weighting across scenes.

\noindent\begin{minipage}{\columnwidth}

\centering
\captionof{table}{\textbf{TransLab surface reconstruction by scene.} PGSR$^*$ and RT-Splatting$^*$ are local reproductions; other baseline scores are quoted from TSGS~\citep{li2025tsgs}.}
\label{tab:labgeoper}
\scriptsize\setlength{\tabcolsep}{1.6pt}\renewcommand{\arraystretch}{0.96}
\resizebox{\linewidth}{!}{%
\begin{tabular}{@{}lrrrrrrrrr@{}}
\toprule
Method & Scene 1 & Scene 2 & Scene 3 & Scene 4 & Scene 5 & Scene 6 & Scene 7 & Scene 8 & Average\\
\midrule
\multicolumn{10}{c}{Chamfer Distance (mm)$\downarrow$}\\
\midrule
SuGaR & 37.17 & 10.34 & 8.01 & 6.48 & 8.55 & 46.15 & 14.01 & 7.44 & 17.27\\
2DGS & 2.50 & 8.63 & 2.24 & \rankthree{2.14} & 4.80 & \rankthree{2.12} & \rankthree{2.12} & 3.86 & 3.55\\
GOF & 4.79 & 10.38 & 2.72 & 3.14 & 10.03 & 18.39 & 9.97 & 4.88 & 8.04\\
PGSR$^*$ & \rankthree{1.83} & 5.31 & 2.47 & 2.99 & 5.13 & 29.49 & 2.37 & 7.27 & 7.11\\
TSGS & \ranktwo{1.68} & \ranktwo{2.42} & \ranktwo{1.57} & \rankone{1.60} & \ranktwo{1.75} & \ranktwo{1.54} & \ranktwo{2.05} & \ranktwo{2.23} & \ranktwo{1.85}\\
RT-Splat.$^*$ & 2.70 & \rankthree{2.71} & \rankthree{2.05} & 2.15 & \rankthree{2.56} & 3.20 & 2.67 & \rankthree{3.10} & \rankthree{2.64}\\
\textbf{ClearForm} & \rankone{1.61} & \rankone{2.06} & \rankone{1.49} & \ranktwo{1.61} & \rankone{1.59} & \rankone{1.31} & \rankone{1.53} & \rankone{2.03} & \rankone{1.65}\\
\midrule
\multicolumn{10}{c}{F1 Score$\uparrow$}\\
\midrule
SuGaR & 0.25 & 0.49 & 0.59 & 0.56 & 0.57 & 0.26 & 0.34 & 0.46 & 0.44\\
2DGS & \rankthree{0.91} & 0.55 & \rankthree{0.90} & \rankthree{0.89} & 0.69 & \rankone{0.98} & \ranktwo{0.97} & 0.84 & 0.84\\
GOF & 0.83 & 0.65 & 0.86 & 0.84 & 0.57 & 0.58 & 0.36 & 0.60 & 0.66\\
PGSR$^*$ & \ranktwo{0.94} & \rankthree{0.79} & 0.87 & 0.85 & \rankthree{0.88} & \ranktwo{0.86} & 0.94 & 0.82 & 0.87\\
TSGS & \rankone{0.97} & \ranktwo{0.91} & \rankone{0.96} & \rankone{0.95} & \rankone{0.97} & \rankone{0.98} & \rankthree{0.96} & \ranktwo{0.94} & \ranktwo{0.95}\\
RT-Splat.$^*$ & 0.90 & \ranktwo{0.91} & \ranktwo{0.92} & \ranktwo{0.91} & \ranktwo{0.93} & \rankthree{0.81} & 0.94 & \rankthree{0.89} & \rankthree{0.90}\\
\textbf{ClearForm} & \rankone{0.97} & \rankone{0.94} & \rankone{0.96} & \rankone{0.95} & \rankone{0.97} & \rankone{0.98} & \rankone{0.99} & \rankone{0.95} & \rankone{0.96}\\
\bottomrule
\end{tabular}}
\end{minipage}\par\medskip
\noindent\begin{minipage}{\columnwidth}

\centering
\captionof{table}{\textbf{TransLab novel-view synthesis by scene.} PGSR$^*$ uses our common-view RGB evaluation; RT-Splatting$^*$ is a local reproduction, and published baseline scores are quoted from TSGS~\citep{li2025tsgs}.}
\label{tab:labnvsper}
\scriptsize\setlength{\tabcolsep}{1.6pt}\renewcommand{\arraystretch}{0.96}
\resizebox{\linewidth}{!}{%
\begin{tabular}{@{}lrrrrrrrrr@{}}
\toprule
Method & Scene 1 & Scene 2 & Scene 3 & Scene 4 & Scene 5 & Scene 6 & Scene 7 & Scene 8 & Average\\
\midrule
\multicolumn{10}{c}{PSNR (dB)$\uparrow$}\\
\midrule
SuGaR & 23.27 & 23.21 & 22.20 & 21.81 & 22.83 & 23.44 & 24.34 & 23.14 & 23.03\\
2DGS & 38.94 & 30.12 & 38.74 & \rankthree{37.68} & 34.60 & 36.07 & 41.45 & 37.68 & 36.91\\
GOF & 24.13 & 24.73 & 23.78 & 23.72 & 23.99 & 24.57 & 23.57 & 24.01 & 24.06\\
PGSR$^*$ & 37.71 & 35.13 & 36.94 & 35.90 & 34.24 & 35.21 & 41.18 & 36.94 & 36.66\\
TSGS & \rankone{41.02} & \rankthree{36.58} & \ranktwo{40.34} & \ranktwo{38.37} & \rankthree{35.41} & \ranktwo{37.37} & \ranktwo{45.72} & \rankthree{37.83} & \ranktwo{39.08}\\
RT-Splat.$^*$ & \rankthree{40.12} & \ranktwo{36.84} & \rankthree{38.93} & 37.64 & \ranktwo{35.52} & \rankthree{36.36} & \rankthree{42.49} & \ranktwo{38.37} & \rankthree{38.28}\\
\textbf{ClearForm} & \ranktwo{40.84} & \rankone{37.47} & \rankone{41.81} & \rankone{39.99} & \rankone{35.90} & \rankone{38.44} & \rankone{46.69} & \rankone{40.00} & \rankone{40.14}\\
\midrule
\multicolumn{10}{c}{SSIM$\uparrow$}\\
\midrule
SuGaR & 0.946 & 0.941 & 0.933 & 0.921 & 0.935 & 0.944 & \rankthree{0.961} & 0.945 & 0.941\\
2DGS & 0.962 & 0.702 & 0.953 & 0.952 & 0.944 & 0.960 & \rankthree{0.961} & 0.955 & 0.924\\
GOF & 0.937 & 0.930 & 0.930 & 0.928 & 0.920 & 0.935 & 0.934 & 0.930 & 0.931\\
PGSR$^*$ & \rankthree{0.994} & \rankthree{0.980} & 0.992 & \rankthree{0.989} & \rankthree{0.977} & \rankthree{0.986} & \ranktwo{0.997} & \rankthree{0.989} & 0.988\\
TSGS & \rankone{0.996} & \rankthree{0.980} & \ranktwo{0.995} & \ranktwo{0.992} & \ranktwo{0.978} & 0.985 & \ranktwo{0.997} & 0.987 & \rankthree{0.989}\\
RT-Splat.$^*$ & \ranktwo{0.995} & \rankone{0.986} & \rankthree{0.994} & \ranktwo{0.992} & \rankone{0.980} & \ranktwo{0.987} & \ranktwo{0.997} & \ranktwo{0.990} & \ranktwo{0.990}\\
\textbf{ClearForm} & \rankone{0.996} & \ranktwo{0.984} & \rankone{0.996} & \rankone{0.994} & \rankone{0.980} & \rankone{0.990} & \rankone{0.998} & \rankone{0.991} & \rankone{0.991}\\
\midrule
\multicolumn{10}{c}{LPIPS$\downarrow$}\\
\midrule
SuGaR & \rankthree{0.114} & 0.133 & 0.144 & 0.157 & 0.130 & 0.110 & \rankthree{0.081} & 0.111 & 0.123\\
2DGS & 0.130 & 0.338 & 0.165 & 0.160 & 0.156 & 0.113 & 0.139 & 0.137 & 0.167\\
GOF & 0.123 & 0.137 & 0.132 & 0.133 & 0.144 & 0.118 & 0.130 & 0.131 & 0.131\\
PGSR$^*$ & \ranktwo{0.010} & \rankthree{0.036} & \rankthree{0.011} & \rankthree{0.012} & \rankthree{0.031} & \rankthree{0.017} & \ranktwo{0.006} & \ranktwo{0.015} & \ranktwo{0.017}\\
TSGS & \ranktwo{0.010} & 0.038 & 0.013 & 0.015 & 0.032 & 0.025 & \ranktwo{0.006} & \rankthree{0.019} & \rankthree{0.020}\\
RT-Splat.$^*$ & \rankone{0.008} & \rankone{0.024} & \ranktwo{0.010} & \ranktwo{0.011} & \rankone{0.026} & \ranktwo{0.015} & \rankone{0.005} & \rankone{0.012} & \rankone{0.014}\\
\textbf{ClearForm} & \rankone{0.008} & \ranktwo{0.028} & \rankone{0.009} & \rankone{0.010} & \ranktwo{0.027} & \rankone{0.013} & \rankone{0.005} & \rankone{0.012} & \rankone{0.014}\\
\bottomrule
\end{tabular}}
\end{minipage}

% Place Appendix E quantitative/qualitative floats together at the top of p. 14.
% Their early declarations preserve the D-to-E text order below them;
% the local caption numbering retains E1 before Section E is reached.
\begin{figure*}[!t]
\begingroup\renewcommand{\thesection}{E}\setcounter{figure}{0}
\centering
\includegraphics[width=0.78\textwidth]{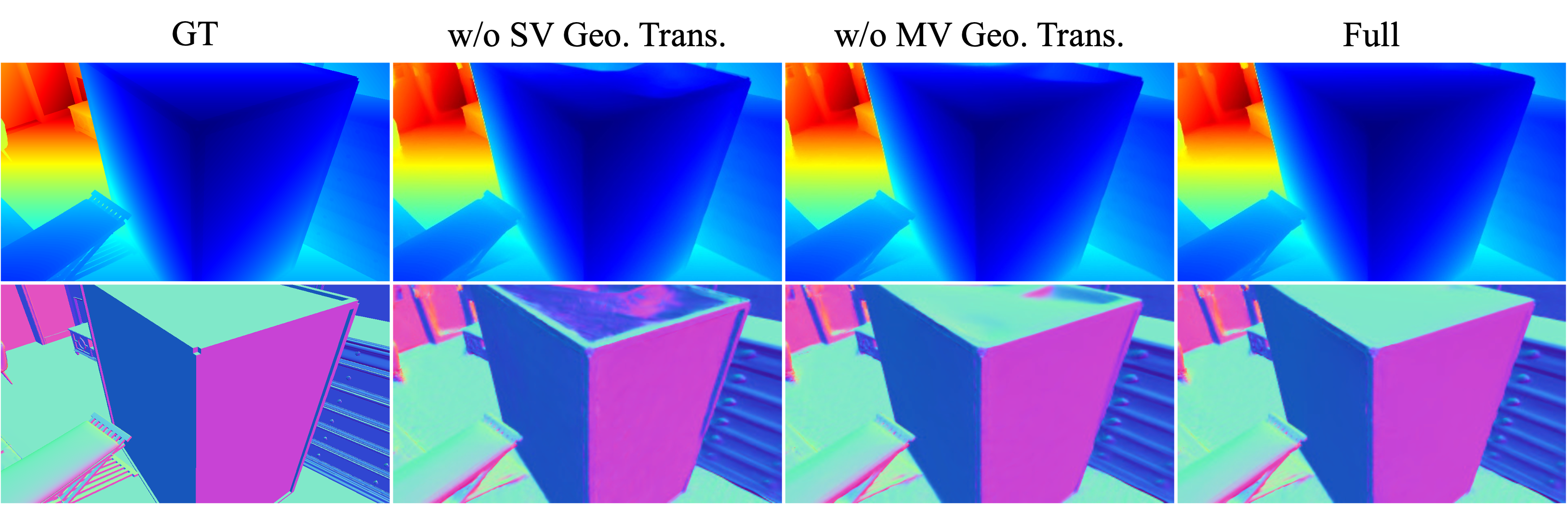}
\caption{\textbf{Transparent-region geometric ablation.} Removing the transparent-region single-view constraint makes the glass top and edges less coherent in the normal map; the multi-view omission has a subtler local effect. Columns show ground truth, the two omissions, and full ClearForm; rows show depth and normals.}
\label{fig:regularizer}
\endgroup
\end{figure*}
\begin{table*}[!t]
\begingroup\renewcommand{\thesection}{E}\setcounter{table}{0}
\centering
\caption{\textbf{Transparent-region regularization on 3D-FRONT-T.} Each omission removes only the transparent-region single-view (SV) or multi-view (MV) contribution; the opaque-region constraints remain active.}
\label{tab:regularizer}
\footnotesize\setlength{\tabcolsep}{3pt}
\resizebox{\textwidth}{!}{%
\begin{tabular}{lrrrrrrrrrrr}
\toprule
\multirow{2}{*}{Configuration} & \multicolumn{3}{c}{Normal} & \multicolumn{3}{c}{Depth} & \multicolumn{2}{c}{Mesh} & \multicolumn{3}{c}{Novel View}\\
\cmidrule(lr){2-4}\cmidrule(lr){5-7}\cmidrule(lr){8-9}\cmidrule(lr){10-12}
& MAE$\downarrow$ & $11.25^\circ\uparrow$ & $22.5^\circ\uparrow$ & AbsRel$\downarrow$ & RMSE$\downarrow$ & $\delta<1.25\uparrow$ & CD (dm)$\downarrow$ & F1$\uparrow$ & PSNR$\uparrow$ & SSIM$\uparrow$ & LPIPS$\downarrow$\\
\midrule
w/o SV Geo. Trans. & 8.33 & 86.54 & 91.54 & 0.027 & 0.07 & 98.31 & 0.28 & 0.922 & 34.08 & 0.956 & 0.053\\
w/o MV Geo. Trans. & 7.44 & 87.37 & 92.24 & 0.025 & 0.07 & 98.38 & 0.26 & 0.930 & 34.14 & 0.956 & 0.053\\
Full & 7.41 & 87.28 & 92.27 & 0.027 & 0.07 & 98.40 & 0.27 & 0.928 & 34.14 & 0.956 & 0.054\\
\bottomrule
\end{tabular}}
\endgroup
\end{table*}
\section{Controlled GACD Diagnostics}
\label{ap:gacd}
We evaluate GT first-hit attribution in 3D-FRONT-T scenes 01--05, independently of DEL. GT transparency masks are eroded by three pixels; view means are averaged per scene, then across scenes.

% Supplementary trained-state/compositing-rule diagnostic; retained for full context.
\begin{table}[!ht]
\centering\small
\caption{\textbf{Checkpoint/rule cross-evaluation (3D-FRONT-T).} Both 30k-iteration checkpoints are evaluated under both geometry-alpha rules. Five-scene means; no DEL or boundary RGB loss.}
\label{tab:gacdreadout}
\setlength{\tabcolsep}{3pt}
\begin{tabular}{lrrrr}
\toprule
Training / alpha rule & MAE$\downarrow$ & AbsRel$\downarrow$ & CD (mm)$\downarrow$ & F1$\uparrow$\\
\midrule
Shared / shared & 21.20 & 0.2164 & 59.25 & 0.794\\
Shared / decoupled & 15.70 & 0.0623 & 52.47 & 0.775\\
Decoupled / shared & 21.36 & 0.2014 & 92.57 & 0.746\\
Decoupled / decoupled & 7.43 & 0.0263 & 27.09 & 0.925\\
\bottomrule
\end{tabular}
\end{table}

For each sampled pixel $p$ in camera $v$, Gaussian $i$ has its own plane-intersection depth, distinct from the blended depth in \cref{eq:plane-depth}. Here we restore the view index: $d_i^v$, $ n_i^v$, and $ r_v(p)$ denote the camera-$v$ instances of $d_i$, $ n_i$, and $ r(p)$ defined above:
\begin{equation}
 z_i^v(p)=\frac{d_i^v}{( n_i^v)^\top r_v(p)}.
 \label{eq:individual-plane-depth}
\end{equation}
After per-view scale alignment, $z_\star^v(p)$ denotes the GT first-surface depth. For each sampled $(v,p)$, we suppress $v$ in the following sets and weights and define $F(p)=\{i:|z_i^v(p)-z_\star^v(p)|<2\,\mathrm{cm}\}$ and $B(p)=\{i:z_i^v(p)>z_\star^v(p)\}$. The latter includes all deeper planes, not a second GT depth layer; the sets can overlap near the GT depth. Under each alpha rule, $A_f(p)=\sum_{i\in F(p)}w_i(p)$ and $A_b(p)=\sum_{i\in B(p)}w_i(p)$. First-hit coverage is the fraction of \emph{sampled} transparent-region pixels with $A_f(p)>0.05$.

Alpha-threshold rejection ($\alpha_{\min}=1/255$) is measured over first-hit pixel--Gaussian pairs active under decoupled alpha. In \cref{sec:gacdexp}, only the compositing rule changes at the common 10,000-iteration state.

\paragraph{Supplementary cross-compositing diagnosis.}
With the shared-trained checkpoint fixed, decoupled geometry alpha improves AbsRel (0.2164 to 0.0623) but lowers F1 (0.794 to 0.775; \cref{tab:gacdreadout}). Under the decoupled rule, decoupled training reduces CD from 52.47 to 27.09\,mm and transparent-region normal MAE from $32.67^\circ$ to $2.62^\circ$; opaque-region MAE changes from $8.71^\circ$ to $8.40^\circ$. Both alpha rule and learned representation matter, but no individual gradient mechanism is isolated.

\section{Additional Ablation Studies}
\label{ap:ablation}

We ablate transparent-region single-view (SV) and multi-view (MV) geometry terms, retaining opaque-region supervision (\cref{tab:regularizer,fig:regularizer}). Removing SV degrades interface-normal accuracy; omitting MV has a smaller effect and slightly improves CD and F1, so a consistent MV benefit is not established.

\section{Additional Real-Scene Results}
\label{ap:real}
\begin{figure*}[t]
\centering\includegraphics[width=0.75\textwidth]{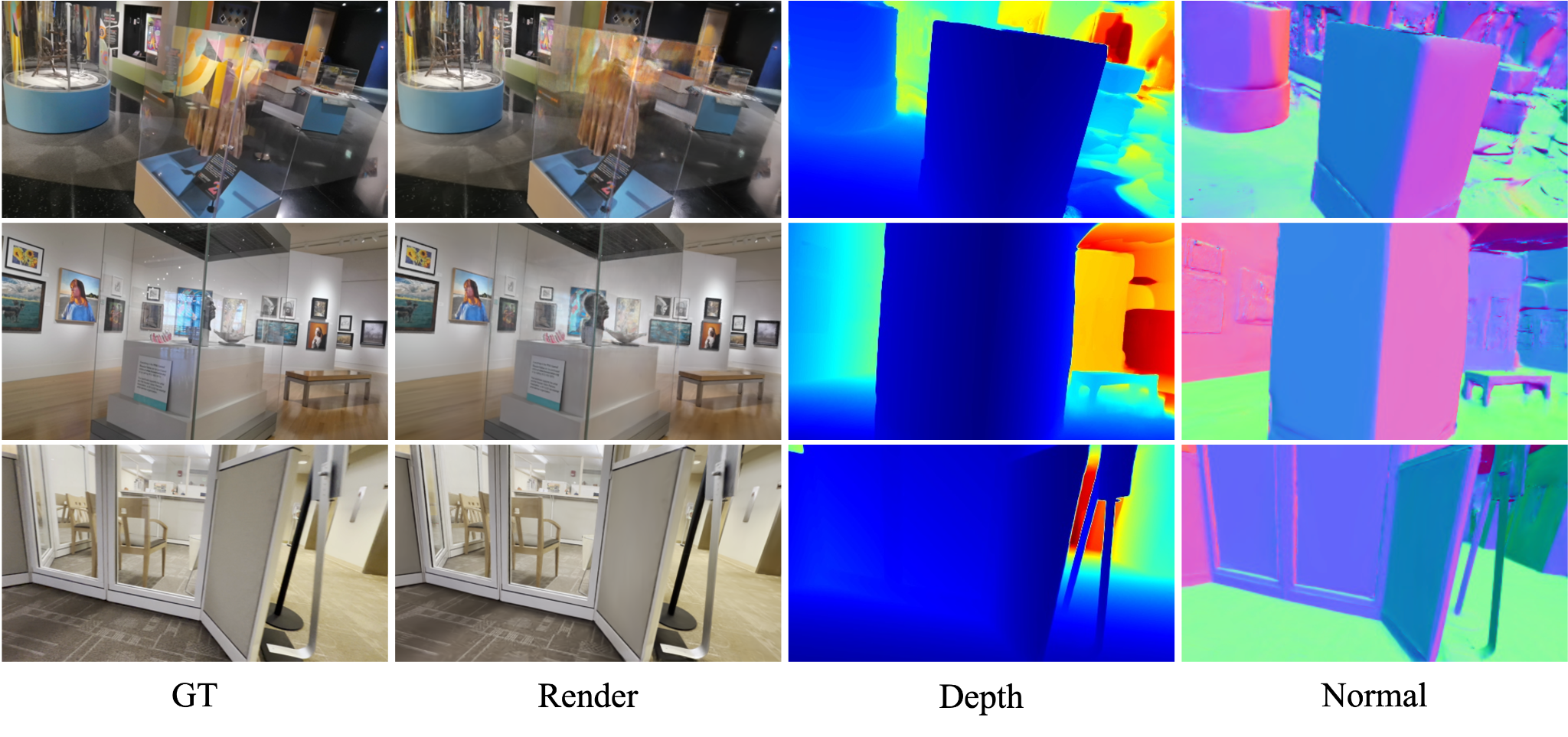}
\caption{\textbf{Additional Ref-DL3DV real-scene results.} Three held-out views illustrate glass display cases and doors. Columns show reference RGB, ClearForm rendering, depth, and normals.}
\label{fig:realadditional}
\end{figure*}

Figure~\ref{fig:realadditional} provides held-out Ref-DL3DV views, evaluated qualitatively because GT geometry is unavailable.

\section{Representative Failure Cases}
\label{ap:visual}
\begin{figure*}[!t]
\centering\includegraphics[width=0.74\textwidth]{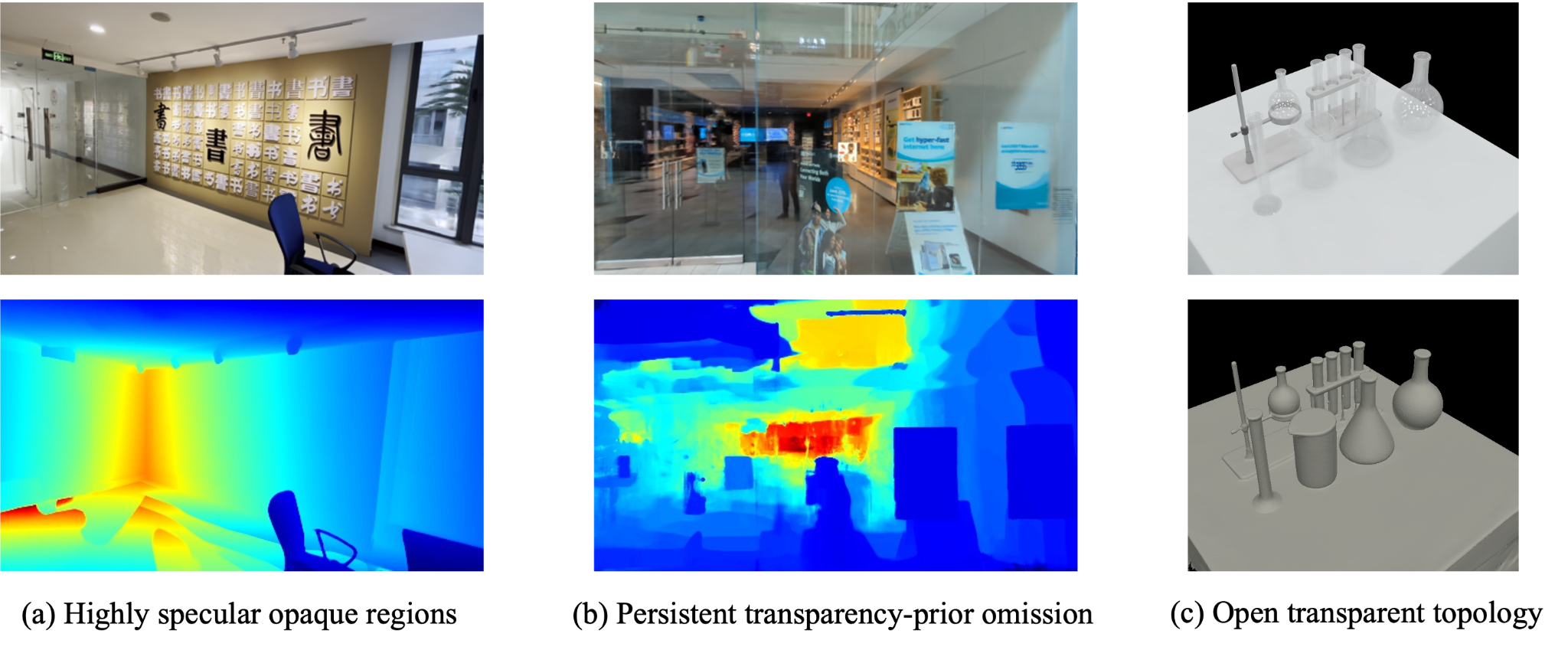}
\caption{\textbf{Representative failure cases.} (a) Strong reflection in an opaque-labeled region biases depth. (b) Missing transparency guidance weakens the glass surface. (c) Single-depth fusion caps an open transparent container.}
\label{fig:limitations}
\end{figure*}

\Cref{fig:limitations} illustrates limitations due to inaccurate transparency priors and single-depth mesh extraction: missed reflective interfaces, incomplete glass geometry, and artificially capped open containers.

% Full-width failure cases remain on the final page.
\clearpage

% END supplement_body.tex

\end{document}